# Imaging radar and LiDAR image translation for 3-DOF extrinsic calibration

Sangwoo Jung[1] · Hyesu Jang[1] · Minwoo Jung[1] · Ayoung Kim[1] · Myung-Hwan Jeon[1]

## Abstract

The integration of sensor data is crucial in the field of robotics to take full advantage of the various sensors employed. One critical aspect of this integration is determining the extrinsic calibration parameters, such as the relative transformation, between each sensor. The use of data fusion between complementary sensors, such as radar and LiDAR, can provide significant benefits, particularly in harsh environments where accurate depth data is required. However, noise included in radar sensor data can make the estimation of extrinsic calibration challenging. To address this issue, we present a novel framework for the extrinsic calibration of radar and LiDAR sensors, utilizing CycleGAN as a method of image-to-image translation. Our proposed method employs translating radar bird-eye-view images into LiDAR-style images to estimate the 3-DOF extrinsic parameters. The use of image registration techniques, as well as deskewing based on sensor odometry and B-spline interpolation, is employed to address the rolling shutter effect commonly present in spinning sensors. Our method demonstrates a notable improvement in extrinsic calibration compared to filter-based methods using the MulRan dataset.

## 1 Introduction

As the field of autonomous driving technology evolves, the significance of sensors that perceive the driving environment increases. In recent times, cameras, LiDAR, and radar sensors have gained widespread adoption in autonomous vehicles. In contrast to traditional vision-based sensors, range-based sensors provide information based on the distance to the objects. LiDAR sensors provide dense 3D laser information in clear weather conditions, while radar sensors furnish approximate point cloud data for the surrounding environment. Due to the nature of the radio waves employed by radar sensors, the quality of data provided by radar sensors is superior to that of LiDAR sensors in adverse weather conditions. Therefore, it is essential to integrate data from both sensors and determine their relationship, referred to as extrinsic calibration, to leverage the benefits of both sensors in vision-degrading conditions fully.

Extensive research has been conducted on extrinsic calibration for both homogeneous sensor combinations [1, 2], and heterogeneous sensor combinations [3–5]. One method of extrinsic calibration is the model-based approach, which aims to optimize a cost function based on a specific target, such as circle-based targets [6], plane fitting [7], and reprojection error [4]. However, the accurate correspondence of data from two sensors is required to perform calibration in this manner, which can be challenging for spinning range sensors due to the rolling shutter effect. This effect occurs when data is acquired while the system is in motion, resulting in sensors providing measurements of point locations closer or farther than the actual distance from the points. This distance error can cause the same point in the acquired data to appear in different areas, leading to errors when attempting to establish correspondence. To address this challenge,

✉ Myung-Hwan Jeon
  myunghwan.jeon@snu.ac.kr

  Sangwoo Jung
  dan0130@snu.ac.kr

  Hyesu Jang
  dortz@snu.ac.kr

  Minwoo Jung
  moonshot@snu.ac.kr

  Ayoung Kim
  ayoungk@snu.ac.kr

1 Department of Mechanical Engineering, Seoul National University, 1 Gwanak-ro Gwanak-Gu, Seoul 08826, Republic of Korea

pre-processing LiDAR and radar pointcloud data is necessary to enable precise extrinsic calibration. In this paper, we reduce the motion distortion by utilizing inertial measurement unit (IMU)-preintegration [8], a technique primarily used in simultaneous localization and mapping (SLAM).

In addition to the challenges posed by the rolling shutter effect, the use of radio waves by radar sensors also leads to significant measurement variations depending on the medium's reflective properties. These characteristics result in measurement data that is heavily contaminated by noise. This paper employs an image-to-image translation method to transform radar images into LiDAR-style images with lower noise levels. To achieve this, we exploit CycleGAN [9] as the image-to-image translation method, which does not require paired images for the training period.

In this study, we present a novel pipeline for the extrinsic calibration between radar and LiDAR sensors, which is based on the translation of radar images into LiDAR-style images using CycleGAN [9]. Additionally, we address the challenge of motion distortion by utilizing inertial measurement unit (IMU)-preintegration [8] to achieve accurate correspondences between the two measurements. Finally, we estimate the 3-degree-of-freedom (DOF) extrinsic parameters using mutual information. The overview of our method is given in Fig. 1 while its contributions are as follows:

1. We propose a novel framework utilizing CycleGAN-based image-to-image translation for estimating 3-degree-of-freedom (DOF) extrinsic parameters between radar and LiDAR sensors.
2. We address the challenge of motion distortion by incorporating the use of IMU-preintegration [8] to enhance the accuracy of 3-DOF extrinsic parameter estimation for heterogeneous sensors which operate at different frequencies.
3. We demonstrate the significant improvement in the performance of 3-DOF extrinsic parameter estimation compared to filter-based methods through experimental evaluation on the MulRan dataset [10].

## 2 Related works

### 2.1 Model-based sensor calibration

Sensor calibration is an essential process for multi-sensor systems. Accurate extrinsic calibration is crucial for precise motion estimation, point cloud mapping, and sensor fusion. Several camera calibration methods [1, 11–13] have been developed for environment perception, highlighting the importance of sensor calibration and the use of vision sensors with geometrical calculations. While camera calibration has played a foundational role in autonomous driving, the need for sensors that provide depth information directly and accurately has increased due to the limitations of cameras that require additional geometry calculations for depth information. With the growing importance of accurate ranging sensors, studies on extrinsic calibration with LiDAR and radar have gained attention. Fremont et al. [6] conducted camera-LiDAR calibration using a circle-based target, while Pusztai and Hajder [3] proposed a camera-LiDAR calibration method using a cardboard box. Jeong et al. [7] developed a calibration method for non-overlapping stereo cameras and LiDAR that leverages static and robust road information. The method includes informative image selection, optimization under edge alignment, plane fitting cost, and normalized information distance.

In the context of LiDAR-only calibration, Muhammad and Lacroix [2] proposed a method for geometrical calibration of a 64-ray LiDAR sensor. Additionally, Atanacio-Jiménez et al. [14] proposed a LiDAR calibration method that utilizes pattern planes, which includes a mathematical model and numerical algorithm for minimizing systematic error in an outdoor environment. Furthermore, Jiao et al. [15] proposed a method for calibrating dual LiDAR sensors that leverages plane extraction and matching. Recently, Das et al. [16] proposed a multi-LiDAR extrinsic calibration algorithm that can be utilized with LiDAR system whose LiDAR field of view FOVs are non-overlapping by exploiting SLAM algorithm and LiDAR semantic features to give correspondences between the LiDAR sensors with different sights.

For radar sensors, Peršić et al. [4] and Domhof et al. [5] proposed methods for the calibration of camera–LiDAR–radar sensor systems. Peršić et al. [4] proposed a special target for data accumulation that includes a two-step optimization procedure, with the first step being the minimization of reprojection error, and the second step being the refinement of high uncertainty parameter subsets. Domhof et al. [5] developed a tool for the camera–LiDAR–radar calibration that provides joint extrinsic calibration results for the three sensors and utilizes a unique target with a trihedral corner reflector. Zhang et al. [17] have proposed 3DRadar2ThermalCalib recently, which aims at extrinsic calibration between a 3D mmWave radar and a thermal camera, utilizing a spherical–trihedral target, encouraging the thermal–radar fusion SLAM works.

Previous studies on multi-modal calibration have relied on using special targets or robust common features, which are not always assumed to be existing around the sensors. In contrast, we propose a novel method for estimating the 3-DOF extrinsic parameters between LiDAR and imaging radar without needing any special targets or robust features.

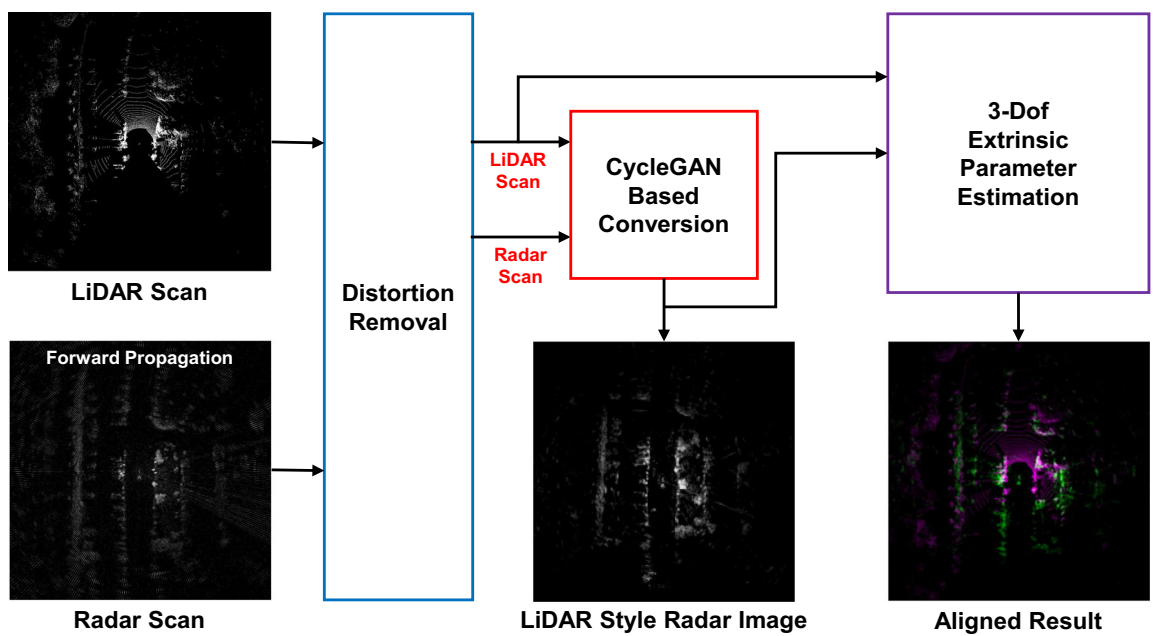

**Fig. 1** The images on the left depict LiDAR and radar bird-eye-view images. Before the CycleGAN-based image conversion (indicated by the red box), distortion removal (indicated by the blue box) is applied to both images. The lower middle image illustrates the result of the CycleGAN-based conversion, which produces a LiDAR-style image with LiDAR data characteristics. Next, 3-DOF extrinsic parameter estimation is conducted using both the original LiDAR image and the generated LiDAR-style image. The lower right image displays the outcome of the alignment, where the original LiDAR image and LiDAR-style radar images are depicted in pink and green, respectively (color figure online)

## 2.2 Learning-based sensor calibration

Due to the rapid growth in the field of deep learning, the field of computer vision has achieved great success. As one of the most important topics in computer vision is extrinsic calibration, the application of deep learning on extrinsic calibration also made significant development. PoseNet [18] utilized convolutional neural network (CNN) on camera pose regression, including location and orientation. Inspired by PoseNet [18], RegNet [19] utilized CNN to solve three conventional calibration steps (feature extraction, feature matching, and global regression) for the extrinsic parameter calibration of multi-modal sensors. CalibNet [20] applied ResNet network structure on the architecture of RegNet [19] while exploiting a self-supervised network, which includes photometric loss and pointcloud loss. Schöller et al. [21] introduced auto-calibration for radar and camera, utilizing CNN-based method to conduct calibration without specific target. Furthermore, a boosting-inspired training algorithm is utilized for robustness.

Meanwhile, as the deformation of the hardware model can affect the long-term accuracy performance of multi-sensor systems, the online calibration method's importance has grown. To achieve higher accuracy of online calibration for multiple heterogeneous sensors, Yuan et al. [22] and Wang et al. [23] proposed RGGNet and SOIC that provide online calibration of camera and LiDAR. RGGNet [22] focused on utilizing a deep generative model for online calibration, while SOIC [23] utilized semantic centroid to solve the calibration problem as a PnP problem. Furthermore, CalibRCNN [24] calculated 6-DOF transformation between 3D LiDAR and 2D camera in real-time by utilizing the LSTM network to extract features and managed both geometric loss and photometric loss to refine the calibration accuracy. Recently, Duy and Yoo [25] proposed Calibration-Net, which provides auto-calibration between LiDAR and camera using cost volume and Convolutional Neural Network, suggesting the possibility of extrinsic calibration between range sensors using depth information.

Deep-learning-based methodologies are one of the practical solutions for the calibration problem between heterogeneous sensors due to their robustness about unpredictable noise that is produced from typical range sensors while acquiring data. Although deep-learning-based extrinsic calibration methods have shown reasonable results based on CNN and RNN networks, two major limitations remain. First, every method targets the extrinsic calibration between the camera and range sensor while using multiple range sensors is emerging. The second limitation is that most of the methods used the KITTI [26] dataset for the train and test dataset, which provides ideally time-synchronized data that can be used as pairs of data for training. If the input data for training are not time synchronized, the result data from other methods will show less quality compared with our method. Unlike existing methods, we propose a LiDAR-radar 3-DOF extrinsic parameter estimation method based on the CycleGAN metric that provides generators that are learned between two different image domains to guarantee cycle consistency, being robust to datasets with unsynchronized images.

### 2.3 Generative Adversarial Networks (GANs)-based image translation

In recent years, the application of deep learning in computer vision has led to significant advancements, particularly in extrinsic calibration. One notable development is the use of GANs for style transfer, which aims to maintain the content of an image while altering its style. The pix2pix model, proposed by Isola et al. [27], utilized conditional GANs to learn a mapping between input and output images. However, this method requires a paired dataset for training. On the other hand, the CycleGAN model proposed by Zhu et al. [9] allows for image-to-image translation tasks without the need for paired data. By introducing cycle consistency losses, Cycle-GAN learns a mapping between a source and target domain in both directions. Other methods, such as DiscoGAN [28] and DualGAN [29], also utilize unpaired data with unsupervised learning but differ in the loss function used. In this work, we utilize the CycleGAN model as a basis for image-to-image translation between radar and LiDAR images due to its ability to work with unpaired data and its verified compatibility with the number of reproduction cases on GitHub.

## 3 Methodology

The proposed method is illustrated in Fig. 2. The first step of our approach is to preprocess the radar images using trajectory data obtained from SLAM methods to correct for motion distortion caused by the spinning nature of radar sensors. Next, we apply a CycleGAN-based image-to-image translation method to convert the radar images into LiDAR-style images with reduced noise levels. Finally, we perform extrinsic calibration between the LiDAR and radar sensors by aligning the LiDAR-style radar images with the real LiDAR data using mutual information (MI)-based registration method and phase correlation-based registration method.

### 3.1 Forward propagation from polar to Cartesian

In the pre-processing stage, we convert radar images from polar coordinates to Cartesian coordinates. Specifically, a radar polar image $I_R^P$ is transformed into a radar Cartesian image $I_R^C$. Two methods are utilized for this transformation process: forward propagation and backward propagation. As shown in Fig. 3, there is a distinct difference between the images generated by these two methods. With forward propagation, points from the polar image are transformed through a one-to-one correspondence, resulting in a sparser Cartesian image as the distance of the point from the center increases. Conversely, backward propagation employs a one-to-many correspondence, yielding a denser image but with higher noise. In this study, we utilize forward propagation for image conversion, as the high noise levels associated with backward propagation may lead to inaccurate 3-DOF extrinsic parameter estimation.

### 3.2 Motion distortion of spinning sensors

As both LiDAR and imaging radar are spinning sensors which acquire a single scene with a rotating module inside the sensor, the position of every point acquired from the sensor is distorted because the full system is moving while the scene is getting acquired. This motion distortion should be handled before the 3-DOF extrinsic parameter estimation.

#### 3.2.1 Motion distortion removal

Before estimating the 3-DOF extrinsic parameter, the trajectory of each sensor can be achieved by SLAM algorithms. We exploit the radar odometry [30] and LiDAR odometry [31] to achieve trajectory of each sensor separately. Additionally, by utilizing high-frequency sensors such as an IMU, the rolling shutter effect can be handled with a high degree of accuracy as the IMU-preintegration can provide the position of the sensor between each SLAM trajectory position.

To correct for any distortions caused by motion, it is essential to merge the points obtained at different time intervals into a single frame. We define the reference frame as the frame where each sensor starts acquiring data, while the frame containing each point is referred to as the point frame. Motion distortion can be corrected by determining the transformation between these two frames. However, even with dense trajectories with IMU-preintegration [8], obtaining a perfectly accurate transformation corresponding to the point frames of all points is challenging due to the IMU cannot provide data at the exact time for every single point because of time synchronization problem between IMU sensor and each ranging sensor. Therefore, based on the estimated trajectory with IMU-preintegration [8], we perform B-spline interpolation [32] to obtain $\mathbf{T}_j^S \in SE(3)$ at a specific time $t_j$. B-spline curve is an interpolation algorithm that receives several points as input and returns the interpolated point at a specific time, whose performance with the range sensors such as LiDAR has been proven as the [33]. With the four nearest trajectory positions about the point in the axis of time,

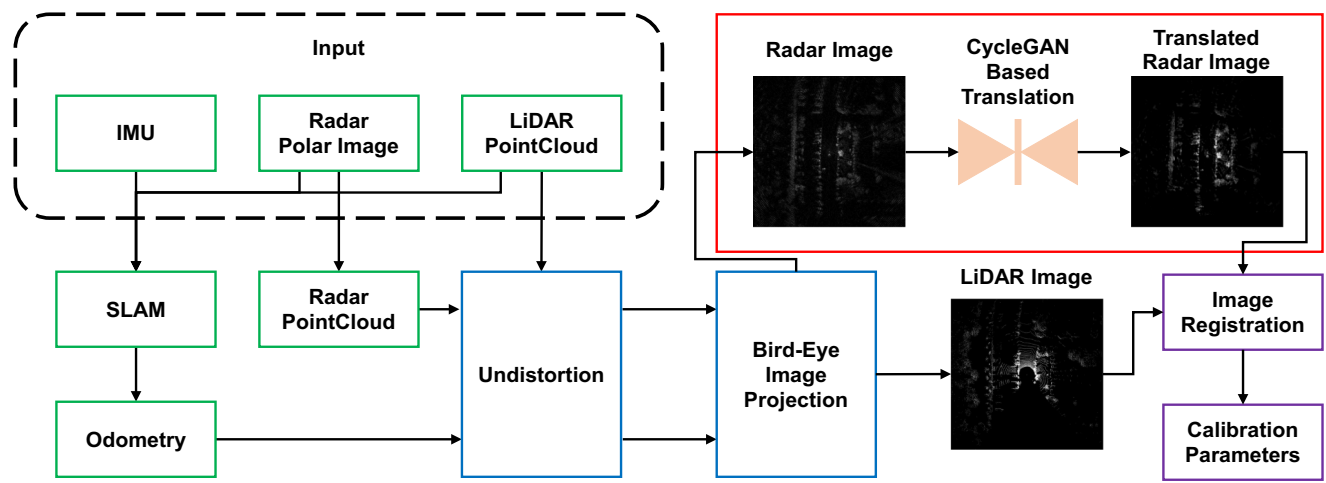

**Fig. 2** This figure illustrates the comprehensive framework of our proposed method. The green box denotes the input data, which is not specified in detail. The blue box represents the data processing component, further explained in Sect. 3.2. In Sect. 3.3, the radar image is transformed into a LiDAR-style image using the CycleGAN-based network, as represented by the red box. Finally, the translated radar and original LiDAR images are utilized to estimate the 3-DOF extrinsic parameters through the purple box (color figure online)

**Fig. 3** **a**, **b** The transformed radar Cartesian image from the original polar image. An image produced by forward propagation includes fewer pixels but less noise than that produced by backward propagation

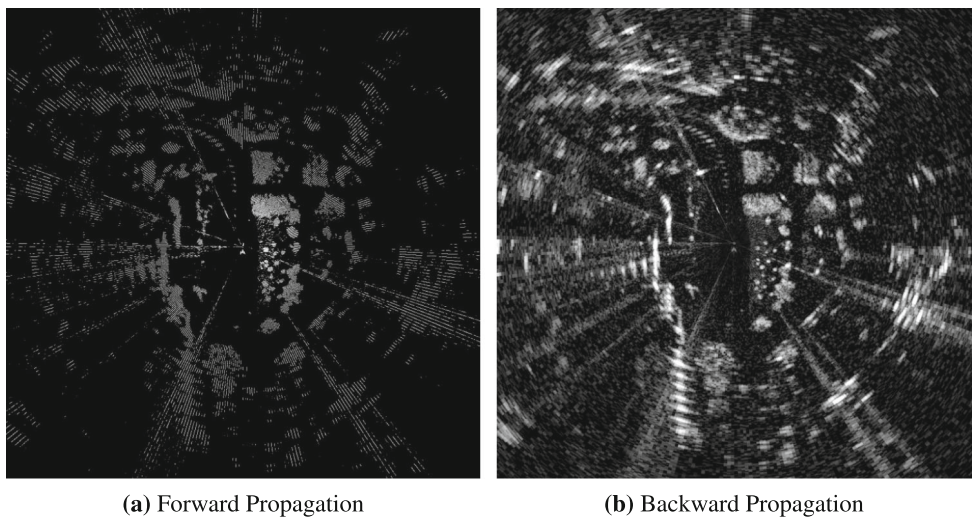

(a) Forward Propagation  (b) Backward Propagation

**Fig. 4** **a** presents the original radar image, with a yellow line indicating the scan baseline and rotation direction. Due to the effect of motion distortion, the beginning and end of the scan are not clearly matched. In contrast, **b** demonstrates the deskewed image, in which pixels are registered in their proper location as a result of motion distortion elimination. The improvement in alignment can be observed through the improved correspondence of orange and blue points, which were acquired at the beginning and end of one scan period, respectively (color figure online)

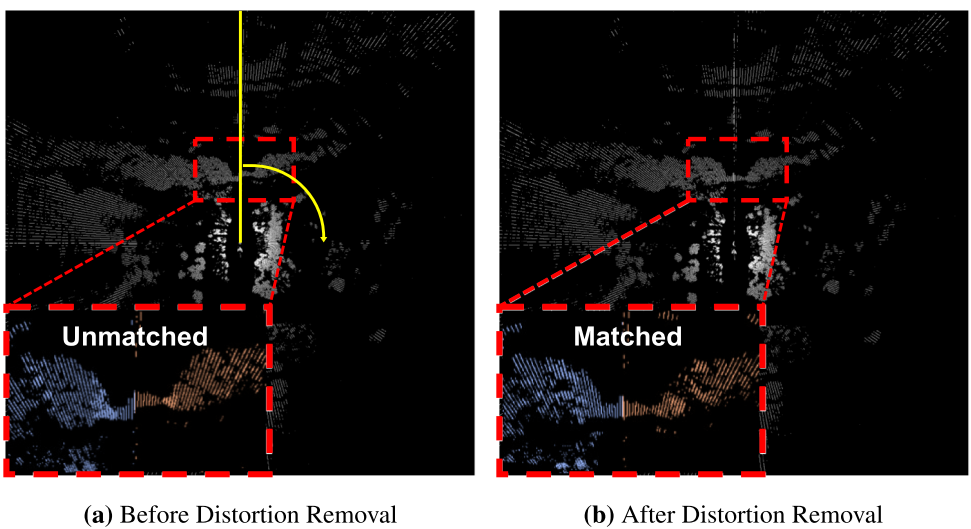

(a) Before Distortion Removal  (b) After Distortion Removal

B-spline interpolation can provide the interpolated position $\mathbf{T}_j^S$.

Finally, refined position of 3D points $p_j^D$ based on the B-spline interpolation are calculated as

$$p_j^D = \left\{\mathbf{T}_i^S\right\}^{-1} \mathbf{T}_j^S \, p_j \tag{1}$$

where $i$ is the time when the data acquire is first started and $j$ is the time when the point is acquired. $\mathbf{T}_i^S$ is position of the system at time $i$ provided from the SLAM trajectory while $\mathbf{T}_j^S$ is position of the system at time $j$ calculated with the B-spline interpolation and IMU-preintegration.

### 3.2.2 LiDAR sensor

In the case of LiDAR, motion distortion is removed from the 3D point cloud through the method outlined in Sect. 3.2.1. The resulting point clouds are then transformed into bird-eye-view images with a pixel size of $p$ and image size of $r$, representing half of the maximum distance.

### 3.2.3 Imaging radar sensor

For the Imaging radar, since the transformed Cartesian radar images are in 2D, we convert these images into a point cloud $P_R$, where each point $p_{R,j}$ is located in the coordinate space $[I_x, I_y, 0]$ while $I_x$ and $I_y$ refers the position of the point in *x*-axis and *y*-axis. We then remove motion distortion through the method outlined in Sect. 3.2.1. The resulting point clouds are subsequently transformed into bird-eye-view images with a pixel size of $p$ and image size of $r$, representing half of the maximum distance.

Figure 4a, b indicates before and after the motion distortion, showing that points from starting and ending of one sweep do not match if the motion distortion is not removed.

### 3.3 Radar to LiDAR image translation

After removing motion distortion, the bird-eye-view LiDAR and radar images are ready for further processing. Due to the radar image and LiDAR image differences in the noise level and data characteristics, direct image registration is unworkable. To address this issue before the image registration process, image translation of radar image into a LiDAR-style image is necessary. The style transfer algorithm, which does not require the perfectly paired dataset, is needed as the exact paired image between radar and LiDAR cannot be provided. To achieve this, we utilize the CycleGAN architecture [9], which does not require a paired dataset for training. The network is trained on bird-eye-view radar images and corresponding LiDAR images.

In the CycleGAN model, there are two generator functions, $G_X : X \rightarrow Y$ and $G_Y : Y \rightarrow X$. Both generators translate images between the X and Y domains, while the X domain includes radar images and the Y domain includes LiDAR images. As such, separate discriminators are used for each generated result: $D_X$ and $D_Y$. $D_X$ discriminates whether the input image belongs to the X domain, and $D_Y$ performs a similar task for the Y domain.

Similar to normal GANs network's loss functions, CycleGAN includes adversarial loss for generators and discriminators. Due to the generators and discriminators being paired, there are two different adversarial loss functions:

$$L_{\text{GAN}}(G, D_Y, X, Y) = \mathbb{E}_{y \in Y}\left[\log D_Y(y)\right] \\ + \mathbb{E}_{x \in X}\left[\log(1-D_Y(G(x)))\right] \tag{2}$$

$$L_{\text{GAN}}(F, D_X, X, Y) = \mathbb{E}_{x \in X}\left[\log D_X(x)\right] \\ + \mathbb{E}_{y \in Y}\left[\log(1-D_X(F(y)))\right] \tag{3}$$

Although the adversarial loss can train the network to generate reasonable images, which should be included in domains *X* and *Y*, the cycle consistency cannot be trained from the basic adversarial loss. To guarantee that the output of mapped images is included in the target domain, the image that has passed both generators should still include the characteristics of the original domain. For our goal, the LiDAR-styled radar image should be included in the radar domain when it goes through the translation of the generator that maps the LiDAR image to the radar image. For this, CycleGAN uses cycle consistency loss:

$$L_{\text{cyc}}(G, F) = \mathbb{E}_{x \in X}\left[||F(G(x)) - x||_1\right] \\ + \mathbb{E}_{y \in Y}\left[||G(F(y)) - y||_1\right] \tag{4}$$

Using the L1 norm, the reconstructed image through two generators is compared with the original image to calculate the cycle consistency loss. This loss function helps to guarantee that the generated LiDAR-style bird-eye-view image would include enough characteristics of the original LiDAR bird-eye-view image.

As the major goal of CycleGAN's generator is to preserve the input image's characteristic of the target domain and delete the original domain's characteristic, the mapped result of $G(y)$ should not change as the input image is already included in the target domain. Therefore, identity loss is also used for CycleGAN, and its equation is like this:

$$L_{\text{identity}}(G, F) = \mathbb{E}_{y \in Y}\left[||G(y) - y||_1\right] \\ + \mathbb{E}_{x \in X}\left[||F(x) - x||_1\right] \tag{5}$$

This identity loss would help the trained network to remain the style of LiDAR included in the original radar image.

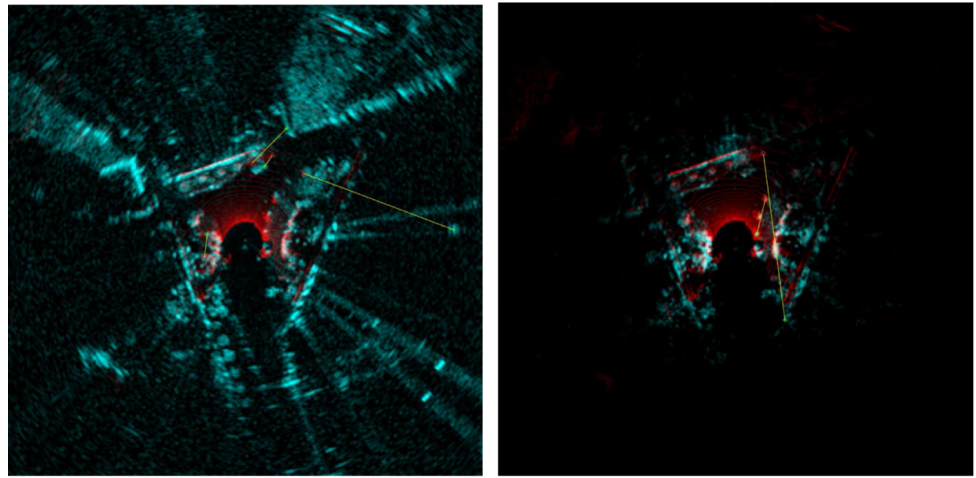

**Fig. 5** Due to the point sparsity, feature-based methods are not able to register LiDAR images to radar images. As the right image, translated radar images with feature extraction methods include less points than the original images. Low correspondences between LiDAR image and both radar and translated radar image are shown

**(a)** Original radar's Feature matching  **(b)** Translated radar's Feature matching

## 3.4 Image registration

In order to estimate the extrinsic parameters between the Imaging radar and LiDAR, we utilize image registration methods. Given that both radar and LiDAR images possess sparse point clouds, traditional feature-based registration methods may not be effective due to the low correspondences between the images, as demonstrated in Fig. 5. Therefore, we employ two direct registration methods. By assuming that only a rigid transformation exists between the two bird-eye-view images, we adopt a phase correlation-based method, which can compute fast and accurate results compared to brute-force matching methods. The cross-correlation value between the two images can be represented mathematically as

$$\frac{1}{n}\sum_{x,y} f(x, y) t(x, y). \tag{6}$$

while $f, t$ are the original and target image signals. In this paper, we adopt intensity as the metric for quantifying image signals, represented by $f$ and $t$ for the original and target images, respectively. The extrinsic parameters are derived from the peak signal that establishes the correlation between the two images. MI-based approach is employed to attain accurate registration results. However, due to the high computational complexity of the MI-based approach as an optimization process, it is utilized for verifying registration accuracy and not for real-time 3-DOF extrinsic parameter estimation.

## 4 Experimental results and discussion

In this section, a thorough explanation of the experimental settings and associated results is provided. The quality of the translated radar images and their registration results with LiDAR images are evaluated and compared with those obtained from the filtered radar images.

## 4.1 Experimental setup

KAIST sequences in the MulRan dataset[10] were utilized for experiments. The MulRan dataset is a dataset that includes multi-modal range sensor data such as radar and LiDAR. Furthermore, IMU sensor data and GPS data are provided from the dataset, making the dataset suitable for the evaluation of methodology. Because the KAIST sequence was acquired at the college campus, the data include many dynamic objects, such as students on the sidewalk or other cars running on the driveways. KAIST_0620 and KAIST_0823 datasets were employed for training, and the KAIST_0902 dataset was used to evaluate 3-DOF extrinsic parameter estimation. Both the training and testing datasets were scaled to a size of $300 \times 300$ pixels, with each image representing a range of $150m \times 150m$, equivalent to 0.5m per pixel. The training dataset consists of 3000 LiDAR and radar bird-eye-view images randomly selected from the KAIST_0620 and KAIST_0823 datasets, with 1500 images from each sequence. Image pairs with an acquisition time difference less than $t_{max}$ were utilized to ensure accurate extrinsic parameter estimation.

We adopt the generative network architecture from Cycle-GAN as described in [9]. The architecture consists of three convolution layers, nine residual blocks as described in [34], two fractionally strided convolutions with $\frac{1}{2}$ stride, and one last convolution layer. Instance normalization [35] was employed with the convolutional layers. To be more specific about the CycleGAN generator, let c7s1 − k be a 7×7 Convolution-InstanceNorm-ReLU layer with k filters and stride of 1. Furthermore, dk denotes a 3×3 Convolution-InstanceNorm-ReLU layer with $k$ filters, including a stride of 2. Rk is a residual block containing two 3×3 convolutional layers, including the same number of filters on both layers, while uk represents a 3×3 fractional-strided-Convolution-InstanceNorm-ReLU layer with $k$ filters and stride of $\frac{1}{2}$. The generator network that has been exploited is like follows: c7s1 − 64, d128, d256, R256, R256, R256, R256, R256, R256, R256, R256, R256, u128, u64, c7s1 − 3. A more detailed description of the generator network is included in Table 1.

**Table 1** Detailed architecture of the CycleGAN generator

| No | Name | Layer (type) | Output Shape | Parameters |
|---|---|---|---|---|
| 1 | c7s1 − 64 | 7×7 Convolution-InstanceNorm-ReLU with 64 filters | [-1, 64, 256, 256] | 3200 |
| 2 | d128 | 3×3 Convolution-InstanceNorm-ReLU with 128 filters | [-1, 128, 128, 128] | 73,856 |
| 3 | d256 | 3×3 Convolution-InstanceNorm-ReLU with 256 filters | [-1, 64, 256, 256] | 295,168 |
| 4 | R256 | ResidualBlock with two 3×3 Convolutions-InstanceNorm-ReLU | [-1, 64, 256, 256] | 1,180,160 |
| 5 | R256 | ResidualBlock with two 3×3 Convolutions-InstanceNorm-ReLU | [-1, 64, 256, 256] | 1,180,160 |
| 6 | R256 | ResidualBlock with two 3×3 Convolutions-InstanceNorm-ReLU | [-1, 64, 256, 256] | 1,180,160 |
| 7 | R256 | ResidualBlock with two 3×3 Convolutions-InstanceNorm-ReLU | [-1, 64, 256, 256] | 1,180,160 |
| 8 | R256 | ResidualBlock with two 3×3 Convolutions-InstanceNorm-ReLU | [-1, 64, 256, 256] | 1,180,160 |
| 9 | R256 | ResidualBlock with two 3×3 Convolutions-InstanceNorm-ReLU | [-1, 64, 256, 256] | 1,180,160 |
| 10 | R256 | ResidualBlock with two 3×3 Convolutions-InstanceNorm-ReLU | [-1, 64, 256, 256] | 1,180,160 |
| 11 | R256 | ResidualBlock with two 3×3 Convolutions-InstanceNorm-ReLU | [-1, 64, 256, 256] | 1,180,160 |
| 12 | R256 | ResidualBlock with two 3×3 Convolutions-InstanceNorm-ReLU | [-1, 64, 256, 256] | 1,180,160 |
| 13 | u128 | 3×3 fractional-strided-Convolution-InstanceNorm-ReLU with 128 filters | [-1, 128, 256, 256] | 295,040 |
| 14 | u64 | 3×3 fractional-strided-Convolution-InstanceNorm-ReLU with 64 filters | [-1, 64, 256, 256] | 73,792 |
| 15 | c7s1 − 3 | 7×7 Convolution-InstanceNorm-ReLU with 3 filters | [-1, 1, 256, 256] | 3137 |

**Table 2** Detailed architecture of the CycleGAN discriminator

| No | Name | Layer (type) | Output Shape | Parameters |
|---|---|---|---|---|
| 1 | C64 | 4×4 Convolution-InstanceNorm-LeakyReLU with 64 filters | [-1, 64, 128, 128] | 1088 |
| 2 | C128 | 4×4 Convolution-InstanceNorm-LeakyReLU with 128 filters | [-1, 128, 64, 64] | 131,200 |
| 3 | C256 | 4×4 Convolution-InstanceNorm-LeakyReLU with 256 filters | [-1, 128, 64, 64] | 524,544 |
| 4 | C512 | 4×4 Convolution-InstanceNorm-LeakyReLU with 512 filters | [-1, 128, 64, 64] | 2097664 |

70×70 Patch GANs [27, 36, 37] were used for the discriminator. For more details about the CycleGAN discriminator, let Ck denote a 4×4 Convolution-InstanceNorm-LeakyReLU layer with k filters and stride 2. To produce a scalar output from the discriminator, convolution is applied. The slope of 0.2 was utilized with Leaky ReLUs, while the discriminator architecture is as follows: C64–C128–C256–C512. A more detailed description of the discriminator network is included in Table 2.

We exploited 0.0002 as the initial learning rate, and Half of the iterations were run with the initial learning rate. Another half of the iterations were run with a linear decay learning rate of zero. The weight for cycle loss is 10.0, while identity loss is set as 2. More specific explanations about the network

**Table 3** Image quality analysis. The highest value is written in **bold**

|  | PSNR | SSIM |
|---|---|---|
| Original radar | 14.4038 | 0.0346 |
| Median-filtered radar | 15.0587 | 0.0580 |
| Gaussian-filtered radar | 14.6781 | 0.0402 |
| CNN-filtered radar | 10.3528 | 0.0125 |
| Translated radar (10 epoch) | 24.0301 | 0.7319 |
| Translated radar (15 epoch) | 24.5550 | 0.7579 |
| Translated radar (20 epoch) | 24.3805 | 0.7606 |
| Translated radar (25 epoch) | 24.4076 | 0.7781 |
| Translated radar (30 epoch) | **24.6909** | **0.7831** |
| Translated radar (35 epoch) | 24.6760 | 0.7819 |
| Translated radar (40 epoch) | 24.4502 | 0.7829 |

structure are well explained in the CycleGAN original paper [9].

### 4.2 Evaluation for translated radar image

In this section, we evaluate the quality of the translated radar images generated using our method. This method aims to

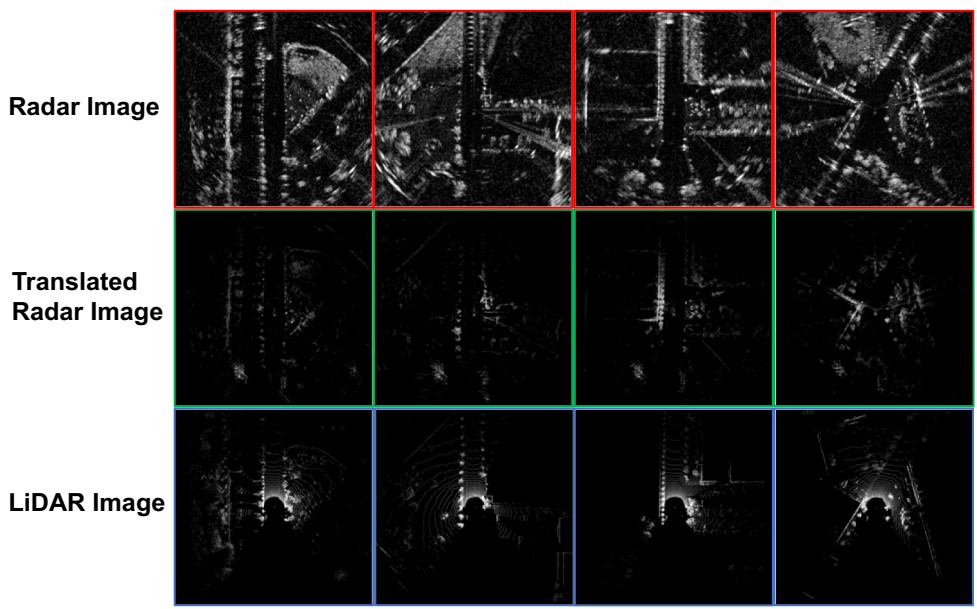

**Fig. 6** Images with red borderlines represent the bird-eye-view radar images generated in the Cartesian coordinate system using back-propagation. Blue-bordered images depict the bird-eye-view LiDAR images, which possess more centered and clean data. Images with green border lines depict the translated radar images generated from the CycleGAN network. These translated images incorporate the data from the radar sensor with reduced noise levels. Most of the ring and radial noise in the red-bordered images is eliminated in the green-bordered images, resulting in a better match with the LiDAR images, as observed intuitively (color figure online)

**Table 4** Extrinsic Calibration Distribution based on MI and phase correlation

| 30 epoch | Mean (Extrinsic parameters) | | | STD (Convergence rate) | | |
|---|---|---|---|---|---|---|
| | x [cm] | y [cm] | $\theta$ [rad] | x [cm] | y [cm] | $\theta$ [rad] |
| CycleGAN-PH | 4.4361 | 40.6015 | −0.0133 | 17.5047 | 24.9289 | 0.1199 |
| CycleGAN-MI | −0.0357 | 56.6839 | −0.0289 | **12.9081** | **22.4413** | 0.2232 |
| CNN-MI | −25.2390 | 13.5681 | −0.0076 | 131.9820 | 125.7971 | 0.0775 |
| Gaussian-MI | −59.3272 | 43.3379 | −0.0086 | 55.3282 | 58.8750 | 0.0095 |
| Median-MI | −63.6261 | 47.5538 | −0.0093 | 54.1358 | 60.6015 | 0.0093 |
| Original-MI | −55.6915 | 39.8573 | −0.0082 | 55.7864 | 56.8529 | 0.0097 |

Smallest standard deviations that leads to the highest convergence rate are expressed with bold numbers
**Original**: Original radar image, **Median**: Median-filtered radar image, **Gaussian**: Gaussian-filtered radar image, **CNN**: CNN-filtered radar image, **CycleGAN**: CycleGAN-based translated radar image, **MI**: mutual information, **PH**: phase correlation translate the radar images into a LiDAR-style representation that captures the properties of LiDAR data. To evaluate these translated radar images, we employ the peak signal-to-noise ratio (PSNR) and structural similarity index measure (SSIM), commonly used metrics for measuring the quality of generated images. PSNR evaluates the information loss in the generated image, while SSIM considers luminance, contrast, and structural information. Both metrics have a higher value as the quality of the generated image improves. As shown in Table 3, the translated radar image at 30 epochs shows the highest PSNR and SSIM values. The translated radar images exhibit exceptional PSNR and SSIM values when compared to the LiDAR images. Based on the definition of PSNR and SSIM, we conclude that the CycleGAN-based image translation method generates images with similar noise levels and characteristics to LiDAR images. Furthermore, to perform the extrinsic calibration of radar and LiDAR, the translated radar image must reflect the characteristics of LiDAR. However, filter-based methods are unsuitable as they are mainly focused on reducing the noise level of the image and have lower PSNR and SSIM values.

Qualitatively, the translated radar images, as shown in Fig. 6, possess a higher degree of centralization of points and a reduced level of ring and radial noise as compared to the original radar images. These characteristics are similar to those of LiDAR images.

### 4.3 Evaluation for image registration

As mentioned above, the translated radar images are registered on LiDAR images using MI and the phase correlation method. The registered results are analyzed both qualitatively and quantitatively to evaluate the performance of the proposed method.

The accuracy of image registration results is evaluated by applying two methods, MI and Phase Correlation, to translated radar images. The results are then presented in Table 4. It is observed that the convergence rate of CycleGAN-based

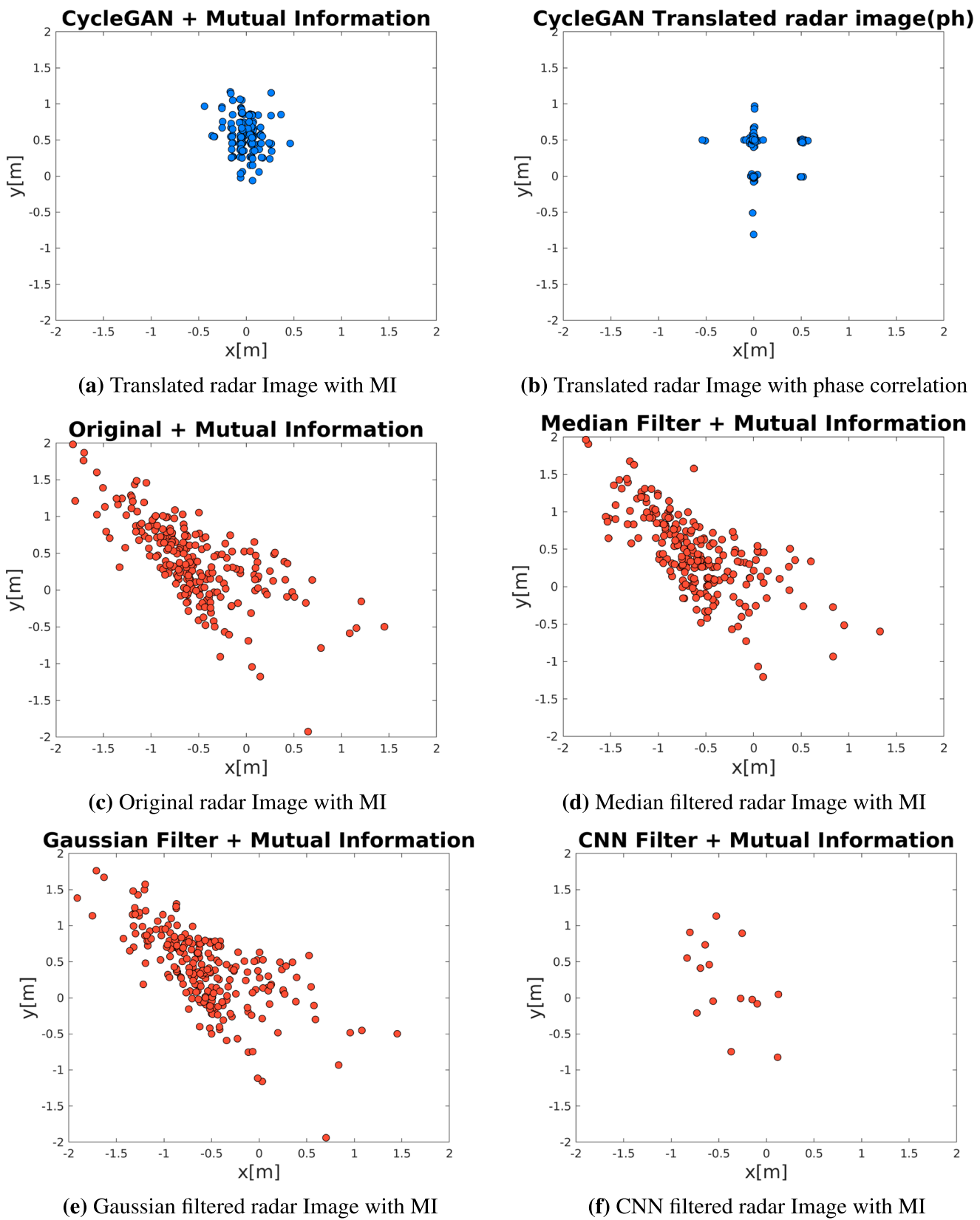

**Fig. 7** **a** and **b** depict the registration results between the CycleGAN-based translated images and LiDAR images using MI and phase correlation, respectively. **c**, **d**, **e**, and **f** depict the registration results of the original radar images and other filter-based radar images with LiDAR images using MI. The points in the images represent the relative translation between the images. After the CycleGAN-based image conversion, we obtained higher precision results than the other methods

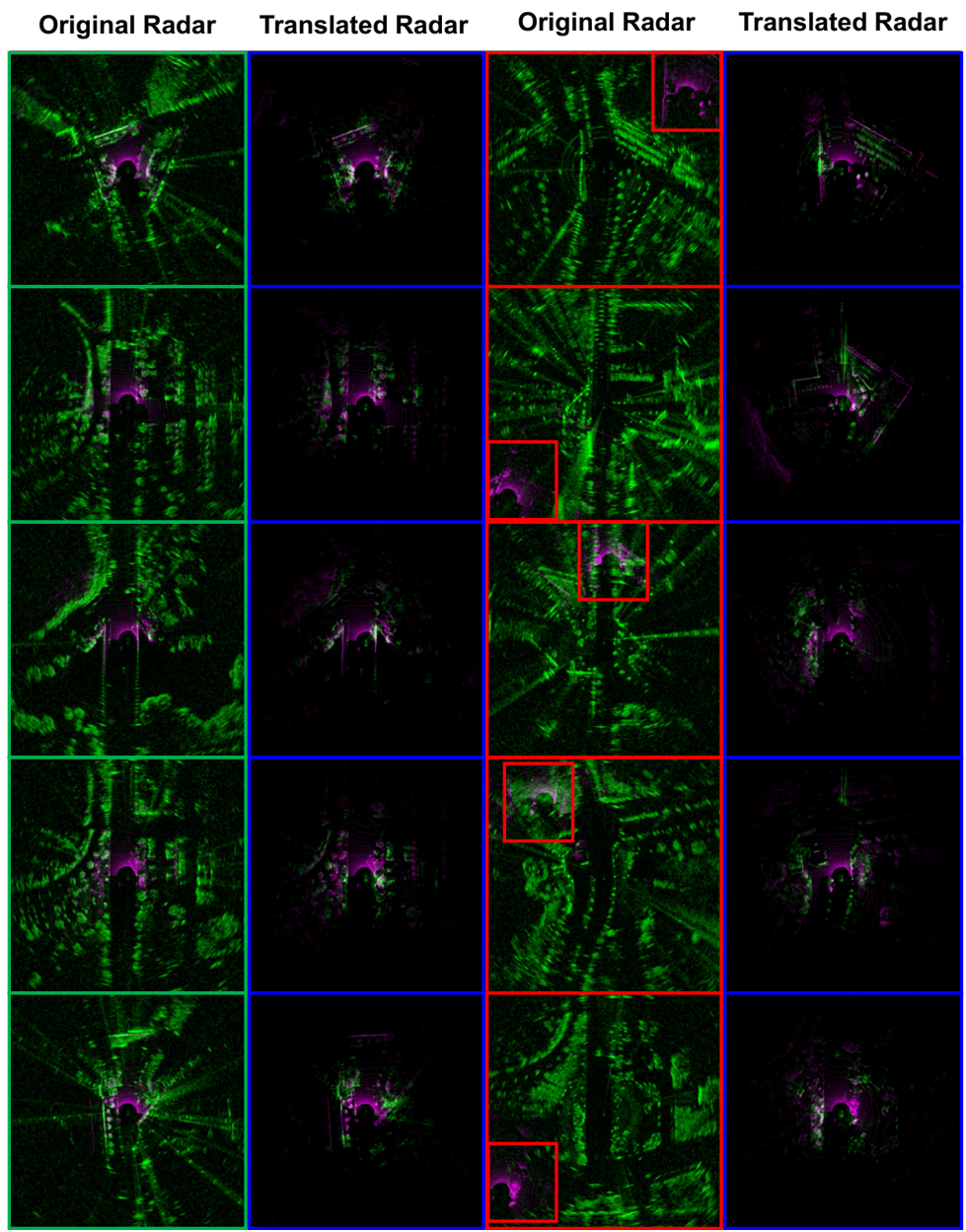

**Fig. 8** The images in the two left columns depict the best-case scenario, where both the original (green borderlines) and translated radar images (blue borderlines) are successfully registered with the corresponding LiDAR images. However, most registration cases involving the original radar images failed, as illustrated by the red borderline images in the third column. In contrast, the CycleGAN-based transformed images in the fourth column (with blue borderlines) demonstrate robust registration results using the same radar data. In each image, the pink points represent the LiDAR data and the green points represent the original radar data or the transformed radar data (color figure online)

generated images is superior to that of filter-based translated images, as evidenced by the low standard deviation values. Additionally, given the sensor configuration of the MulRan [10] dataset, where the radar and LiDAR sensors are arranged in a row, it is found that the MI-based image registration with the LiDAR-style image has achieved convergence to the correct value.

The registration results based on each method are further illustrated in Fig. 7, which confirms the degree of convergence of the image registration results. Figure 7c, 7d, and 7e demonstrates that the Median filter and Gaussian filter were unsuccessful in removing a high level of noise from the input radar image, as the scattered points retained similar shapes.

Figure 7e illustrates the failure of the CNN filter to generate an appropriate LiDAR-style image from the input radar image, as a majority of the points diverged by more than 5 m. Figure 7b presents the results of the phase-correlation-based image registration with the translated radar images, which exhibit a more stable convergence than other filter-based methods. The results of MI-based image registration with the translated radar images, as depicted in Fig. 7b, demonstrate that this combination yields the most stable convergence and exceptional performance. Based on the qualitative analysis, it is concluded that MI-based image registration with CycleGAN-based translated images is more suitable for the extrinsic calibration of radar and LiDAR systems compared with traditional filter-based methods.

The qualitative results of the image registration, as depicted in Fig. 8, demonstrate the effectiveness of the MI-based image registration approach. The fourth column of Fig. 8 illustrates the stable registration results obtained using translated radar images. In contrast, the third column of Fig. 8 depicts the high ring and radial noise present in the original radar images, which consistently leads to unsuccessful registration on the LiDAR images. An interesting point of the LiDAR-style-translated images is that compared with the original radar images; translated images do not include the penetrated points that are generated due to the characteristic of radio waves, which leads to that the CycleGAN-based translation not only reduces the noise level but also removes the radar style points that is not possible to be acquired with LiDAR sensors. These findings conclude that the translated radar images exhibit a more stable registration on LiDAR images than the original radar images for two reasons: *(1)* less ring and radial noise level, *(2)* less characteristic of radar point.

## 5 Conclusion

This paper proposed a novel pipeline for the extrinsic calibration between radar and LiDAR sensors. The pipeline includes the adjustment of images from each sensor to remove motion distortion through deskewing utilizing the SLAM odometry, IMU-preintegration, and B-spline interpolation. Additionally, the pipeline employs CycleGAN to translate radar images to LiDAR-style images, effectively preserving significant information. The final step involves the application of phase correlation and MI to obtain approximate extrinsic parameters. The proposed pipeline was experimentally validated, showing improved accuracy in the extrinsic calibration. The results of this study provide a foundation for 2D image-based range sensor extrinsic calibration.

In future, we would like to adapt CycleGAN-based radar to LiDAR image translation in other fields, such as place recognition, which is one of the important topics from the SLAM community. Furthermore, as our work has removed some of the discrepancies between radar sensor data and LiDAR sensor data, we consider it would be possible to apply various SLAM methods that could not have been exploited for radar sensors until now.

**Acknowledgements** This study is a part of the research project, "Development of core machinery technologies for autonomous operation and manufacturing (NK242H)," which has been supported by a grant from National Research Council of Science & Technology under the R&D Program of Ministry of Science, ICT and Future Planning. This work was supported by the National Research Foundation of Korea (NRF) grant funded by the Korea government (MSIT) (No. RS-2023-00241758).

**Funding** Open Access funding enabled and organized by Seoul National University.